\documentclass[letterpaper, 10 pt, conference]{ieeeconf}

\IEEEoverridecommandlockouts   
\usepackage[hyphens]{url}  
\usepackage{graphicx} 
\usepackage[figurename=Figure]{caption} 

\usepackage{amsmath}       
\usepackage{amssymb}
  {

  }
\usepackage{subcaption}
\usepackage{tcolorbox}
\usepackage{calrsfs}
\usepackage{color}
\definecolor{myred}{rgb}{0.8,0,0}
\definecolor{mygreen}{rgb}{0,0.6,0}
\definecolor{myblue}{rgb}{0,0,0.7}

\usepackage{algorithm}
\usepackage[noend]{algpseudocode}

\def\bbbr{{\rm I\!R}}

\newcommand{\R}{{\bbbr}{}}

\title{\LARGE \bf
Divide \& Conquer Imitation Learning
}

\author{Alexandre Chenu$^{1}$,  Nicolas Perrin-Gilbert$^{1}$ and Olivier Sigaud$^{1}$
\thanks{$^{1}$Sorbonne Université, CNRS, Institut des Systèmes Intelligents et de Robotique, ISIR F-75005 Paris, France
        {\tt\small chenu@isir.upmc.fr}}%
}

\begin{document}
\maketitle
\begin{abstract}
When cast into the Deep Reinforcement Learning framework, many robotics tasks require solving a long horizon and sparse reward problem, where learning algorithms struggle. In such context, Imitation Learning (IL) can be a powerful approach to bootstrap the learning process.
However, most IL methods require several expert demonstrations which can be prohibitively difficult to acquire. Only a handful of IL algorithms have shown efficiency in the context of an extreme low expert data regime where a single expert demonstration is available. In this paper, we present a novel algorithm designed to imitate complex robotic tasks from the states of an expert trajectory.
Based on a sequential inductive bias, our method divides the complex task into smaller skills. The skills are learned into a goal-conditioned policy that is able to solve each skill individually and chain skills to solve the entire task. We show that our method imitates a non-holonomic navigation task and scales to a complex simulated robotic manipulation task with very high sample efficiency.
\end{abstract}

\section{Introduction}
\label{sec:intro}

Deep Reinforcement Learning (DRL) has been successful in solving complex simulated (\cite{andrychowicz2017hindsight, akkaya2019solving, ecoffet2021first}) and physical robotic control problems \cite{sermanet2018time}. However, even in simulation, DRL is still limited when applied to complex tasks including sparse reward signals \cite{matheron2019problem}, long control-time horizons and critical states \cite{kumar2021should}. In this context, Imitation Learning (IL) is a fruitful alternative to failing DRL algorithms. In IL, a number of expert demonstrations are used to guide the learning process so that the behavior of an agent matches that of an expert. 

\begin{figure}[ht!]
     \centering
     \includegraphics[width = 0.95 \hsize]{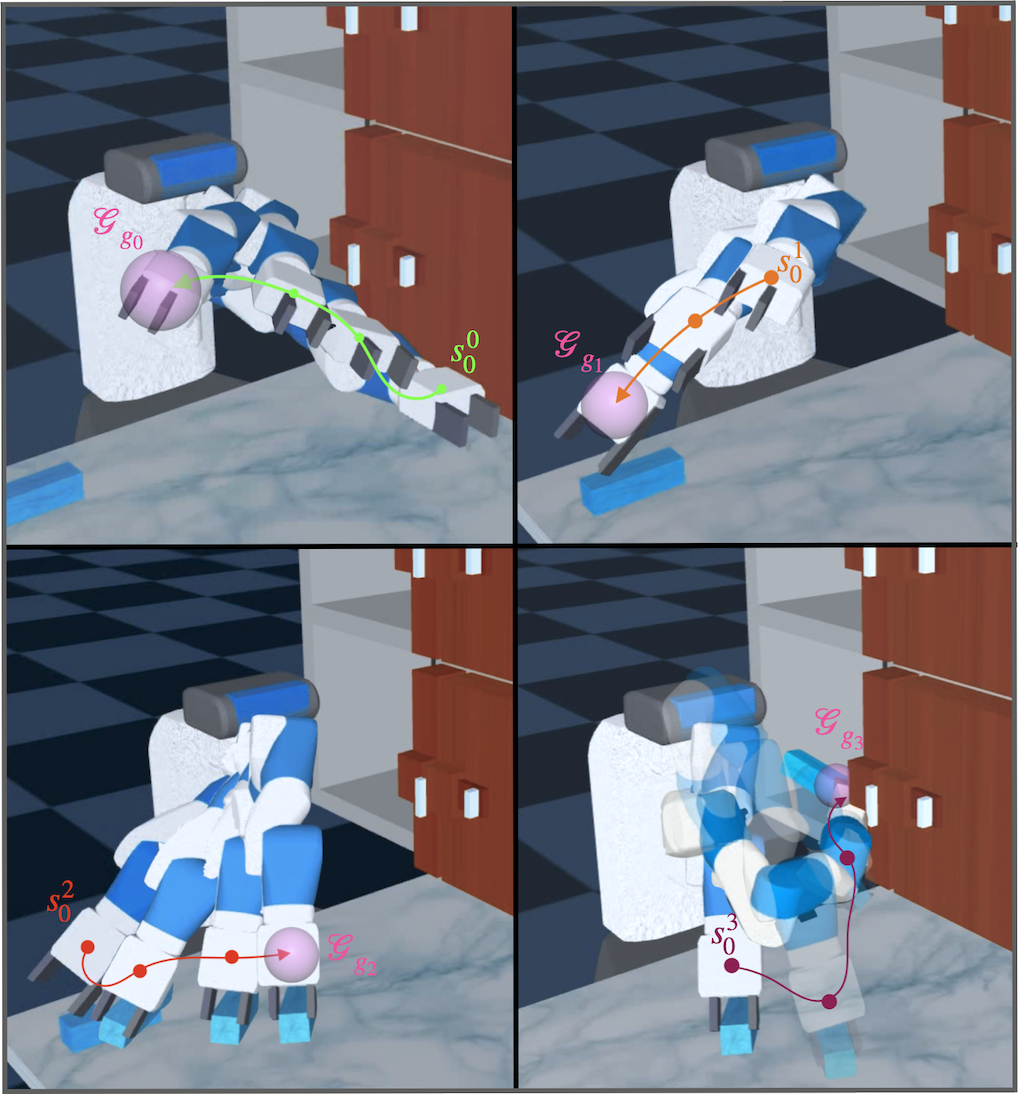}
    \caption{Chaining of four skills learned using DCIL in a simulated object transportation task. The pink sphere represents the success zones defined in the goal space. The goal space contains the Cartesian positions of the end effector and a Boolean indicating if the object is grasped or not.}
    \label{fig:GGI_concept_fetch} 
\end{figure}

However, most imitation learning algorithms require a large set of expert demonstrations which can be hard to acquire, particularly in the context of long-horizon problems. In this context, a few methods strive to design an IL algorithm that can work with a single demonstration. Among these methods, the Go-explore approach (\cite{ecoffet2019go, ecoffet2021first} relies on a strategy called Backplay \cite{resnick2018backplay, salimans2018backplay}) which learns a single controller by starting further and further away from the final point. As it needs to learn and play many longer and longer trajectories, this approach suffers from sample inefficiency. Another recent approach that can handle single demonstrations is PWIL \cite{dadashi2020primal}, which uses offline learning to define an episodic reward function based on the demonstration, and performs IL without formulating it as an adversarial learning problem, contrary to most recent approaches. This improves the efficiency and stability of the IL process, but this still resorts to performing DRL on a long horizon task, which limits the potential gain in sample efficiency.

In this paper, we present Divide \& Conquer Imitation Learning (DCIL), a DRL-based IL algorithm relying on a sequential inductive bias to solve long-horizon imitation
tasks using a single demonstration. As the name implies, DCIL divides the complex task into skills. The skills are learned into a goal-conditioned policy (GCP) that is able to solve each skill individually and chain skills to solve the entire task. In complex problems, it may be necessary for the goal space to have a lower dimensionality than the state space, which means that skills may lead to states that do not match the expert demonstration. As a result, the chaining of the skills becomes challenging, and we address this issue by introducing, for each skill, a chaining reward bonus that depends on a value function learned over the next skill.  
We first evaluate our approach in a toy Dubins maze environment where the dynamics of the controlled system is constrained, and show that our chaining mechanism plays a crucial role in ensuring the success of the method, resulting in a sample efficiency that is several orders of magnitude better than that of Backplay and PWIL. We then turn to a more challenging Fetch environment where an object has to be grasped and put into a drawer with a simulated robotic arm, and demonstrate an even greater gain in sample efficiency compared to Backplay, the method used by Go-Explore on this benchmark.

\section{Related work}

IL is usually transformed into an optimization problem whose objective is to reproduce the behavior of an expert. It can be done directly in Behavioral Cloning \cite{pomerleau1991efficient}, which
relies on regression to learn a policy that mimics the actions of the expert. A more indirect approach is Inverse Reinforcement Learning (IRL) \cite{russell1998learning}, which consists in estimating an unknown reward function from demonstrations of an expert considered optimal and training a policy using the learned reward function. 

These approaches are severely limited by the necessity to measure the actions of the expert, and by their typical need for many demonstrations. More specifically, with few demonstrations, BC tends to suffer from compounding error caused by covariate shift \cite{pmlr-v9-ross10a}. 
In the case of IRL, it can be difficult to extract a reward from a unique demonstration. For instance, popular adversarial methods for IRL (\cite{Ho2016, kostrikov18, ding19, henderson2018optiongan}) rely on a generator-discriminator architecture that may become unstable if the discriminator is not trained on sufficiently many samples from expert trajectories.

There are contexts in which demonstrations are rare or difficult to generate, but only a few of the deep learning-based methods are capable of producing good results in this low data regime. In our experimental validation, we  mainly consider two of them: PWIL and Backplay.


\subsection{IL from a single demonstration}

\subsubsection{PWIL} 
Primal Wasserstein Imitation Learning (PWIL) \cite{dadashi2020primal} is a recent IRL method that minimizes a greedy version of the Wasserstein distance between the state-action distributions of the agent and the expert. The Wasserstein distance presents several good properties that have been proficiently used in the Deep Learning community \cite{arjovsky2017wasserstein}. PWIL solves an occupancy matching problem between an agent and the demonstration without relying on adversarial training, which makes it more stable than adversarial methods. More specifically, it defines an episodic reward function based on the demonstration, and performs IL by maximizing this reward without introducing an inner minimization problem as adversarial approaches do. PWIL achieves strong performances in complex simulated robotics tasks like humanoid locomotion using one single demonstration. 

\subsubsection{Backplay}
The Backplay algorithm (\cite{resnick2018backplay}, \cite{salimans2018backplay}), is an approach explicitly designed for IL from a single demonstration. It has been used in the robustification phase of the first version of the Go-Explore algorithm \cite{ecoffet2019go} to achieve state-of-the-art results on the challenging Atari benchmark Montezuma's revenge and in the Fetch problem that we tackle in Section~\ref{sec:scaling}. 
In Backplay, the objective is to reach the final state of the expert demonstration. The RL agent is initialized close to the rewarding state and the starting state is progressively moved backward along the demonstration if it is successful enough at reaching the desired state. Backplay can be seen as a curriculum for RL approaches in the context of sparse reward and long-horizon control \cite{florensa2017reverse}. 

\subsection{Skill-chaining}

In this paper, we propose to address the single demonstration imitation problem by transforming a demonstration into a sequence of RL tasks. 
This divide \& conquer type of strategy is a common way to solve a complex RL problem by learning a set of policies on simpler tasks and chaining them to solve the global task \cite{konidaris2010}. For example, this principle is applied by the Backplay-Chain-Skill part of the Play-Backplay-Chain-Skill (PBCS) algorithm \cite{matheron2020pbcs}. The Backplay algorithm is used to learn a set of skills backward from the final state of a single demonstration obtained using a planning algorithm. However, in PBCS, the agent must reach the neighborhood of a precise state to transit from one skill to the next. 
In high-dimensional states, the constraint of reaching a sequence of precise states quickly becomes a very hard learning problem, and as a result PBCS struggles to scale to complex robotic tasks.

There are similar approaches in the recent skill-chaining literature (\cite{bagaria2019option, bagaria2021robustly}), in which skills are formalized using the option framework (\cite{sutton1999between, precup2000temporal}). An option is composed of an initial set of states in which this option can be activated, a termination function which decides if the option should be terminated given the current state, and the intra-option policy which controls the agent at a single time-step scale to execute the option. After completing an option, the agent uses an inter-option policy to decide which option should be applied depending on its current state. 
In Deep Skill Chaining (DSC) \cite{bagaria2019option}, for a given goal, a first option is learned to reliably reach it from a nearby region. Then, iteratively, a chain of options is created to reach the goal from further states. The goal of each option is to trigger the initiation condition of the next one, and the phase of construction of the initiation classifiers requires various successful runs randomly obtained via exploration or RL. This framework has not been applied to imitation, but it is possible that a modified version could address IL. However, with a single demonstration and a complex problem, the initiation conditions could end up being very small and precise, in a similar way to PBCS, with the same difficulty to scale to high-dimensional problems. Some adversarial approaches rely on the framework of options to efficiently perform IL (e.g. \cite{henderson2018optiongan}), but as other AIL-based methods, they tend to fail in the low expert data regime that we consider.

In our method, we consider a goal space as a low-dimensional projection of the state space. Instead of targeting the neighborhood of a precise state to complete a skill as in PBCS, we aim at the neighborhood of a low-dimensional goal. Moreover, skills are not performed by independent policies as in the option framework. Instead, we learn a single goal-conditioned policy able to perform different skills depending on the goal it is conditioned on. Finally, skills are not trained independently. Along the chain of skill, the policy is trained in order to complete a skill by reaching states that are compatible with the execution of the following skill.

\section{Background}
\label{sec:background}

In our proposed approach, we extract a sequence of targets from the demonstration, and rely on the formalism of Goal-Conditioned Reinforcement Learning (GCRL) to learn a unique policy able to reach the consecutive targets in order.

\subsection{Goal-conditioned Reinforcement Learning}
\label{sec:GCRL}
A DRL problem is described by a state space $\mathcal{S}$, an action space $\mathcal{A}$, an unknown reward function $R:\mathcal{S}\times \mathcal{A}\rightarrow \R$, an unknown transition probability $p(s_{t+1}|s_{t},a_{t})$, and potentially a distribution of initial states. In a finite horizon setting, an episode has a maximum length of $T_{max}$ control steps. A GCRL problem extends the RL formalism to a multiple goal setting where the reward function $R:\mathcal{S}\times \mathcal{A}\times\mathcal{G}\rightarrow \R$ depends on the goal that is considered. In addition to sampling an initial state, a goal is sampled at the beginning of each episode using a goal distribution $\rho_{\mathcal{G}}$.

The objective in GCRL is to obtain a goal-conditioned policy (GCP) \cite{schaul2015universal}  $\pi(a|s,g)$ that maximizes the expected cumulative rewards $\mathop{\mathbb{E}}[\sum_{t=0}^{T_{max}}R(s_{t},g)]$ for a goal $g$.

\subsection{Distance-based sparse reward}

In most DCRL settings, the reward signal is sparse as the goal-conditioned agent only receives a reward for achieving the desired goal $g$. 

\begin{figure}[t!]
    \centering
    \includegraphics[width = 0.8\hsize]{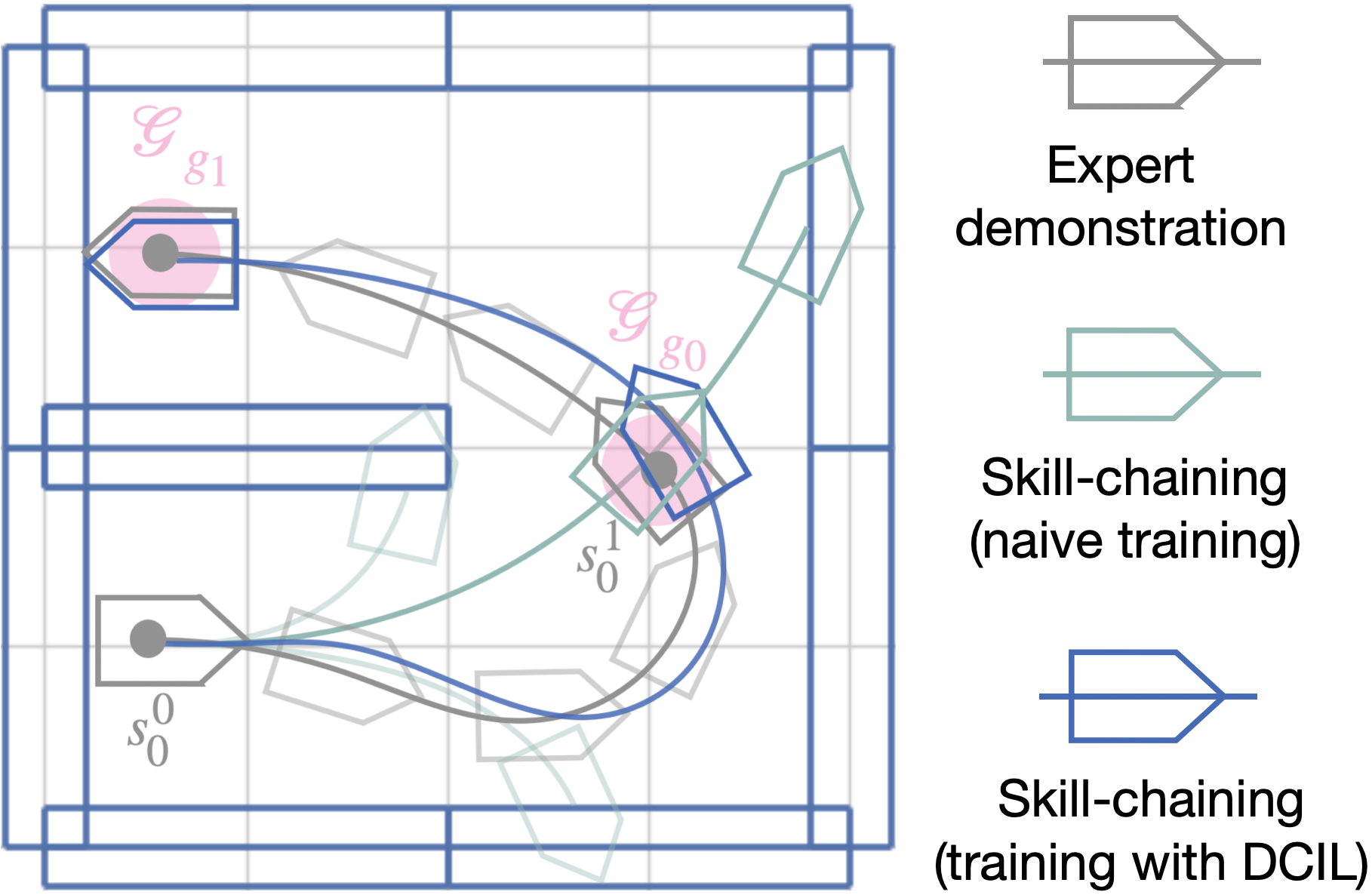}
     \hfill
    \caption{In this example, an expert demonstrates how to navigate a Dubins car (grey trajectory) in a simple 2D maze. The demonstration is split into a set of skills. Here, the two skills consists in reaching $g_0$ and $g_1$, two x-y positions represented as $\mathcal{G}_{g_0}$ and $\mathcal{G}_{g_1}$, whereas states are three-dimensional and include the orientation of the car. If the GCP is trained naively, the agent could solve the first skill by reaching states with an invalid orientation for the next skill (green trajectory). DCIL helps the agent to complete skills by reaching success states with a valid orientation to successfully chain the skills (blue trajectory).}
    \label{fig:DCIL_method}
\end{figure}

In our method, we consider a common version of GCRL where goals represent low dimensional projections of states. A state is projected to a goal according to a mapping $p_{\mathcal{G}}:\mathcal{S}\rightarrow \mathcal{G}$ associated to the definition of the goal space. To achieve a goal, the agent must transit to any state $s\in \mathcal{S}$ that can be mapped to a goal $g_{s}=p_{\mathcal{G}}(s)\in \mathcal{G}$ within a distance less than $\epsilon_{success}$ from $g$, $\epsilon_{success}$ being an environment-dependent hyper-parameter. Those {\em success states} form the success state set $\mathcal{S}_{g}$ associated with goal $g$ and their corresponding goals constitute its success goal set $\mathcal{G}_{g}$. We use the common L2-norm to compute the distance between two goals but other norms can be considered. 
Note that reaching a goal corresponding to a low-dimensional projection of a state does not fully condition the state that the agent is in. This can be very problematic when chaining two skills as illustrated in ~\figurename~\ref{fig:DCIL_method} and discussed in Section~\ref{sec:success_chain}.

The environment-agnostic reward function is defined as:

\begin{equation}
    R(s,g) = \left\{\begin{matrix}
1 \text{ if } s\in \mathcal{S}_{g} \\ 
0 \text{ otherwise.}
\end{matrix}\right.
\end{equation}

To assess how much reward can be expected by following policy $\pi$ conditioned on a goal $g$ from state $s$, we use a goal-conditioned value function $V^{\pi}$ defined as the expected sum of future rewards, given $s$ and $g$:

\begin{equation}
    V^{\pi}(s,g) = \mathop{\mathbb{E}}[\sum_{t=0}^{T_{max}}R(s_{t},g) | s,\pi(.,g)].
\end{equation}

This value function is the central tool used to compute the chaining reward bonus $R_{\text{bonus}}$ (see Section~ \ref{sec:success_chain}). 

\subsection{Relabelling}

Exploring the state space in the context of sparse reward can be challenging even for modern deep RL algorithms (\cite{houthooft2016vime,bellemare2016unifying,burda2018exploration}). To simplify exploration in the GGI framework, GCRL agents often use the Hindsight Experience Replay (HER) relabelling technique \cite{andrychowicz2017hindsight}. If the agent fails to reach the goal it is conditioned on, HER relabels the transitions of the episode by replacing the goal initially intended with the goal it accidentally achieved.

\section{Methods}
\label{sec:methods}

The Divide \& Conquer Imitation Learning (DCIL) algorithm is designed to solve what can be called a Goal-Guided Imitation (GGI) problem. In a GGI problem, instead of imitating the whole expert demonstration, we rely on the \textit{divide \& conquer} paradigm and divide the imitation problem into learning a sequence of goal-conditioned chainable skills. This implies a small loss of generality as it relies on the assumption that demonstrations can be decomposed into a sequence of goal-conditioned tasks. Arguably, this assumption is often true, especially in a robotic context, and as we show in Section~\ref{sec:experiments}, the GGI approach can significantly accelerate the IL process.

\subsection{Goal-Guided Imitation}
\label{sec:ggi}
The Goal-Guided Imitation (GGI) framework can be formulated as a variant of GCRL. A set of $N_{skills}$ goal-based skills are extracted from a single expert demonstration $\tau_{e} = \{s_{0}^{e}, s_{1}^{e},...,s_{N}^{e}\}$ and the objective is to obtain a GCP that is able to complete each skill sequentially.
From this trajectory $\tau_{e}$, we derive skills by extracting a set of goals $(g_{i})_{i\in[0,N_{skills}]} \in \mathcal{G}^{N_{skills}}$. Each skill $K_{i}$ is defined by its goal $g_{i}$. To avoid any ambiguity, we call skill-goals the goals $g_{i}$ associated with skill $K_{i}$.
The objective in the GGI framework is to obtain a GCP that is able to reach skill-goals $g_{i+1}, g_{i+2},...$ after completing skill $K_{i}$. This GCP is then used to chain the successive skills in order to reach the final state $s_{N}^{e}$ of the expert trajectory. 
Unlike in GCRL, in the GGI framework, 
$\rho_{g}$ is a distribution of skill-goals only, as only skill-goals may be sampled to condition the GCP.


\subsection{DCIL hypotheses}
\label{sec:hypothesis}

We formulate three main hypotheses in DCIL. We assume a weak form of \textit{reset-anywhere}, that expert actions are not provided and that a definition of the goal space is given. 

\subsubsection{Reset}
Training the GCP in DCIL (see Section~\ref{sec:learning_tasks}) assumes that the agent can be reset in some selected states of the expert demonstration. A similar form of reset is assumed in Backplay which requires that the agent can be reset in each demonstrated state. Stronger forms of reset such as the assumption that the agent can be reset in uniformly sampled states (\textit{reset-anywhere}) have also been considered in the GCRL literature \cite{nasiriany2019planning}. 
PWIL is based on the more classical assumption of a unique reset, and BC does not require any reset at all.

\subsubsection{No expert actions}
Similarly to Backplay, the imitation in DCIL is solely based on the expert trajectory in the state space. Learning from states only is crucial when the expert actions are difficult to collect (e.g. human demonstrations). On the contrary, both PWIL and BC require state-action demonstrations. 

\subsubsection{Goal-space definition}
As a GCRL-based method, DCIL requires a definition of the goal space and the corresponding mapping from the state space to the goal space. No such assumption is made in Backplay, PWIL or BC as none of them uses a GCP.

\subsection{The Divide \& Conquer Imitation Learning algorithm}

In DCIL, we extract skills goals and initial states from the expert trajectory (Section~\ref{sec:div_traj}). The GCP is then trained to perform each skill using a DRL algorithm. Training for a skill boils down to starting in the associated initial state and completing a local rollout to reach the skill-goal (Section~\ref{sec:learning_tasks}). While the agent learns a skill, it is encouraged to complete it by reaching states that are compatible with the execution of the next ones (Section \ref{sec:success_chain}). Finally, the agent can recover the expert behavior by chaining the skills sequentially (Section~\ref{sec:expert_behavior}). These different stages of DCIL are detailed in the four next sections and summarized in Algorithms~\ref{algo:GCP_training} and ~\ref{algo:SAC_update}.

\subsubsection{Extracting skills from the expert trajectory}
\label{sec:div_traj}

To transform the expert trajectory into teachable skills, we project it in the goal space and divide it into $N_{skill}$ sub-trajectories $(\tau_{i})_{i\in[0,N_{skill}]}$ of equal arc lengths $\epsilon_{\text{dist}}$. For each sub-trajectory, we extract one tuple $(s^{i}_{0}, g_{i}, T^{i}_{max})$ that we associate to a skill, where $s^{i}_{0}$ corresponds to the demonstrated state that resulted in the initial goal of the sub-trajectory, $g_{i}$ the initial goal of the next sub-trajectory and $T^{i}_{max} = \beta |\tau_{i}|$, where $|\tau_{i}|$ is the length of the sub-trajectory in time steps, and $\beta > 1$ is a predefined coefficient\footnote{In our experiments, we use $\beta$ = 1.25.} used to facilitate exploration while learning the skill (see Section~\ref{sec:learning_tasks}).
The initial goal of the next sub-trajectory $g_{i}$ constitutes the skill-goal. The initial state $s^{i}_{0}$ and the length $T^{i}_{max}$ are used to learn the skill.

\begin{algorithm}[t!]
\caption{DCIL - GCP training}
\label{algo:GCP_training}
\begin{algorithmic}
    \State \textbf{Input:} $\pi_{\phi}, Q_{\theta}, Q_{\bar{\theta}}$ 
    \Comment{actor, critic and target critic networks}
    \State $B\leftarrow []$ 
    \Comment{replay-buffer}
    \For{$n=1:N_{episode}$}
        \State $ (s_{0}^{n}, T^{n}_{max}, g_{n}) \leftarrow select\_skill()$
        \Comment{\textbf{Step 1}}
        \State $s_{t} \leftarrow \text{env.reset}(s_{0}^{n}$)
        \Comment{\textbf{Step 2}}
        \State $t \leftarrow 0$
        \While{not done}
            \State $a_{t} \leftarrow \pi_{\phi}(s_{t}|g_{n})$
            \State $s_{t+1} \leftarrow env.step(a_{t})$ 
            \State $r_{t} \leftarrow 0$
            \If{$|p_{\mathcal{G}}(s_{t+1})-g_{n}|_{2} \leq \epsilon_{success}$}
                \State $success, done \leftarrow True, True$
            \Else
                \State $success, done \leftarrow False, False$
            \EndIf
            \If{$t \geq  T^{n}_{max}$}
                \State $done \leftarrow True$
            \EndIf
            \State $B \leftarrow B + [(s_{t}, a_{t}, s_{t+1}, r_{t}, g_{n}, done, success)]$
            \State $t \leftarrow t + 1$
            \If{success}
                \Comment{\textbf{Step 3}}
                \State $success, done \leftarrow False, False$
                \State $(\_, T^{n}_{max}, g_{n}) \leftarrow \text{next\_skill}(g_{n})$
                \Comment{overshoot}
                \State $t \leftarrow 0$
            \EndIf
            \State $\text{SAC\_update}(\pi_{\phi}, Q_{\theta}, Q_{\bar{\theta}},B)$ 
            \Comment{Algo~\ref{algo:SAC_update}}
        \EndWhile
    \EndFor
\end{algorithmic}
\end{algorithm}

\subsubsection{Learning the skills}
\label{sec:learning_tasks}

To train the GCP on the different skills, DCIL runs a three-step loop. 

\paragraph{Step 1} 
DCIL selects a skill $K_{i}=(s^{i}_{0}, g_{i}, T^{i}_{max})$ to train on (function select\_skill in Algorithm~\ref{algo:GCP_training}) and resets the environment in $s^{i}_{0}$.
Note that the selection of skills is biased towards skills with a low ratio of successful rollouts over the total number of trials for these skills. We implemented such distribution using a fitness proportionate selection \cite{Blickle1996selection} where the fitness corresponds to the inverse of this ratio. 

\paragraph{Step 2} 
DCIL conditions the GCP on $g_{i}$ and the agent performs a rollout to complete the skill which is interrupted either if the agent reaches $\mathcal{S}_{g_{i}}$ or if skill length ${T^{i}}_{max}$ is exceeded. 

\paragraph{Step 3} 
When the agent successfully completes skill $K_{i}$, DCIL applies an overshoot mechanism (see Section~\ref{sec:success_chain}) and returns to Step 2. Otherwise, the complete loop is repeated.

The rollouts are saved in a unique replay buffer. For each interaction with the environment, the saved transitions are sampled in a batch to perform a SAC update \cite{haarnoja2018soft} of the critic and actor networks including the chaining reward bonus (see Section~\ref{sec:success_chain}). In a sampled batch of transitions, half of the transitions are relabelled using HER. 
This three-step loop is summarized in Algorithm~\ref{algo:GCP_training}.

\begin{algorithm}[t!]
\caption{modified SAC update (+ HER + Chaining Reward Bonus)}
\label{algo:SAC_update}
\begin{algorithmic}
    \State \textbf{Input:} $\pi_{\phi}, Q_{\theta}, Q_{\bar{\theta}}, B$
    \Comment{actor, critic and target critic networks}
    \State $batch_{HER} \leftarrow HER(B)$ 
    \Comment{HER relabelling}
    \State $batch \leftarrow B$ 
    \Comment{no HER relabelling}
    \For{$(s_{t}^{k}, a_{t}^{k}, s_{t+1}^{k}, g^{k}, r^{k}, done, success)$ in $batch$}
        \If{success}
        \Comment{Chaining reward bonus}
            \State $(\_, \_, {g^{k}}') \leftarrow \text{next\_skill}(g^{k})$
            \State $r^{k} \leftarrow 1 + Q_{\bar{\theta}}(s_{t+1}^{k},\pi_{\phi}(s_{t+1}^{k}, {g^{k}}'), {g^{k}}')$
        \EndIf
    \EndFor
    \State $batch \leftarrow batch + batch_{HER}$
    \For{$gradient\_step=1:N_{gradient\_step}$}
    \Comment{see \cite{haarnoja2018soft}}
        \State{ $\pi_{\phi}, Q_{\theta}, Q_{\bar{\theta}} \leftarrow gradient\_step(\pi_{\phi}, Q_{\theta}, Q_{\bar{\theta}}, batch)$}
    \EndFor
    \State \textbf{Output:} $\pi_{\phi}, Q_{\theta}, Q_{\bar{\theta}}$
\end{algorithmic}
\end{algorithm}

\subsubsection{Ensuring successful skill chaining}
\label{sec:success_chain}
To ensure that skills can be chained, we use an \textit{overshoot mechanism} and a \textit{chaining reward bonus}. 

\begin{figure}[t!]
     \centering
     \includegraphics[width = \hsize]{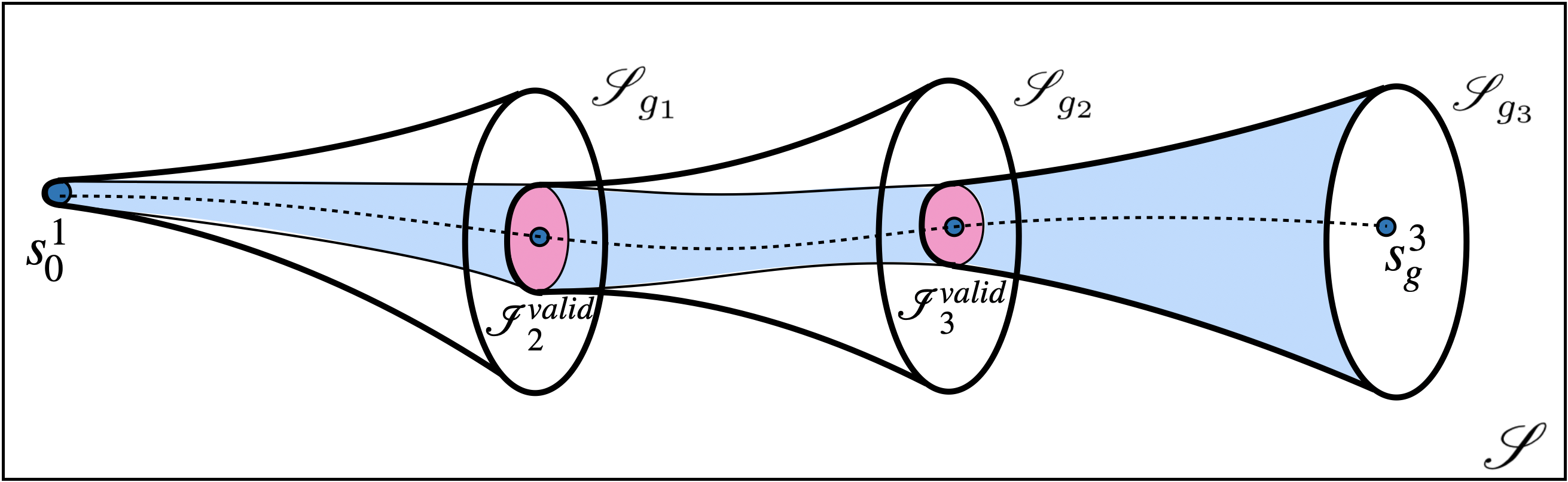}
    \caption{To retrieve the expert behavior (dotted line), the agent has to perform each skill (represented as funnels) sequentially. To successfully chain the skills the agent must transit between their set of valid initial states $\mathcal{I}_{i}^{valid}$ (pink disks) via the blue-shaded path and avoid solving a skill by reaching the other success states in $\mathcal{S}_{g_{i}} \setminus \mathcal{I}_{i}^{valid}$.}
    \label{fig:GGI_chain_concept} 
\end{figure}

\paragraph{Overshoot}
Successfully chaining the skills requires that the agent completes any first skill by reaching states from which it is able to perform the next ones. When the agent completes a skill, it can reach either \textit{valid initial states} in $\mathcal{I}_{i}^{valid}$ from which it will successfully perform the next skills or \textit{invalid initial states} from which it will not (see Figure~\ref{fig:GGI_chain_concept}). 
During a training rollout for a skill, if the agent reaches a success state, the overshoot mechanism immediately conditions the GCP on the skill-goal of the next skill (function next\_skill in Algorithm~\ref{algo:GCP_training}). The agent instantly starts a new rollout for this next skill from its current state. 

With these overshoot rollouts, the agent can learn how to perform the next skills while starting from other initial states than the ones extracted from the demonstration. As the agent progresses at performing skills from those different initial states, some previously invalid initial states become valid. So the purpose of the overshoot mechanism is to make $\mathcal{I}_{valid}$ grow. However, in complex environments (e.g. under-actuated and non-holonomic environments) not all invalid initial states can become valid. Therefore, another mechanism is necessary to help the agent to complete skills by only reaching valid starting states.

\paragraph{Chaining reward bonus}
To facilitate skill completion by only reaching valid initial states for the next skills, we add a chaining reward bonus to the sparse reward received by the agent each time it successfully completes a skill. The chaining reward bonus is defined as the goal-conditioned value function $V^{\pi}(.,g_{i+1})$ of the agent, conditioned on the skill-goal of the next skill (see Algorithm~\ref{algo:SAC_update} for additional information on goal-conditioned value computation). Therefore, the modified reward function is defined as:

\begin{equation}
    \bar{R}(s,g_{i}) = \left\{\begin{matrix}
1 + V^{\pi}(s,g_{i+1}) \text{ if } s\in\mathcal{S}_{g_{i}}\\ 
0\ \text{otherwise.}
\end{matrix}\right.
\end{equation}

The idea behind this bonus is that valid initial states should have a higher value than invalid ones. Indeed, by training the value function $V^{\pi}(.,g_{i+1})$ with transitions extracted from successful episodes and successful overshoots from valid initial states, the rewards from skills $K_{i+1}, K_{i+2}, ...$ are propagated backward up to valid initial states. 

In complex environments, if $\mathcal{I}_{i}^{valid}$ only contains a few isolated valid initial states (or even only the demonstrated state $s_{0}^{i}$), the chaining bonus reward may be difficult to propagate along the skills. A well-balanced entropy-regularized soft update of SAC, using either a hand-tuned entropy coefficient or an adaptive one, forces the agent to explore diverse trajectories to reach $\mathcal{S}_{g_i}$ \cite{eysenbach2021maximum}. It helps the agent to eventually find a path towards valid initial state and propagate the chaining bonus reward.

\subsubsection{Retrieving the expert behavior}
\label{sec:expert_behavior}
To reproduce the expert behavior, we reset the environment in the initial state of the expert demonstration. We condition the GCP on the skill-goal $g_{0}$ of the first skill $K_{0}$. After completing this first skill, the agent is conditioned on $g_{1}$ and solves $K_{1}$. This process is repeated for every skill $K_{i}$ until the final skill is completed.


\section{Experiments}
\label{sec:experiments}

In this section, we introduce the experimental setup used to evaluate DCIL (Section~\ref{sec:xp_setup}), we present an ablation study of the two main components of DCIL (Section~\ref{sec:ablation_study}) and we compare DCIL to three baselines: BC, Backplay and PWIL (Section~\ref{sec:baselines_dubins}). The code of DCIL based on Stable Baselines 3 \cite{raffin2021stable} is provided here: \url{https://github.com/AlexandreChenu/dcil}. 

\subsection{Experimental setup}
\label{sec:xp_setup}

We evaluate DCIL in two environments: the \textit{Dubins Maze} environment that we introduce and the \textit{Fetch} environment presented in \cite{ecoffet2021first}. 

\subsubsection{Dubins Maze}
In the Dubins Maze environment, the agent controls a Dubins car \cite{dubins1957curves} in a 2D maze. The state $s=(x,y,\theta)\in X\times Y \times \Theta$ where $(x,y)$ are the coordinates of the center of the Dubins car in the 2D maze and $\theta$ is its orientation. The forward speed of the vehicle is constant with value 0.5 and the agent only controls the variation $\dot{\theta}$ of the orientation of the car. 
The goal space associated with this environment is $X\times Y$. In the absence of a desired orientation in the skill-goal conditioning the GCP, the agent can easily reach a success state for a given skill with an orientation that is invalid for the next skills. 
Demonstrations for this environment are obtained using the Rapidly-Exploring Random Trees algorithm \cite{lavalle1998rapidly}.

\subsubsection{Fetch}
We also evaluate DCIL in the simulated grasping task for a 8 degrees-of-freedom deterministic robot manipulator. This environment was presented in the First Return then Explore paper \cite{ecoffet2021first}. The objective is to grasp an object initialized in a fixed position on a table and put it on a shelf. The state is a 604-dimensional vector which contains the Cartesian and angular positions and the velocity of each element in the environment (robot, object, shelf, doors...) as well as the contact Boolean evaluated for each pair of elements. On the opposite, the goal only corresponds to the concatenation of the 3D coordinates of the end-effector of the manipulator with a Boolean indicating whether the object is grasped or not. Therefore, the agent may complete a skill prior to the contact with the object with an invalid state (e.g. with an orientation or a velocity that prevents grasping). Demonstrations are collected using the exploration phase of the Go-Explore algorithm \cite{ecoffet2019go}.

\subsubsection{Baselines}
We compare DCIL to three IL methods. The first baseline is a naive BC method using a single demonstration. The two others are state-of-the-art methods that are proficient in the context of imitation from a single demonstration: Backplay (\cite{salimans2018backplay, resnick2018backplay}) and PWIL \cite{dadashi2020primal}. A comparison of the different key assumptions required by each algorithm is detailed in Section~\ref{sec:hypothesis}. For PWIL, we used the implementation provided by the authors with the same hyper-parameter. For Backplay, we re-implemented it to evaluate it in the Dubins Maze and extracted the results obtained for the Fetch environment in \cite{ecoffet2021first}.

\begin{figure}[!t]
     \centering
     \includegraphics[width = 0.9 \hsize]{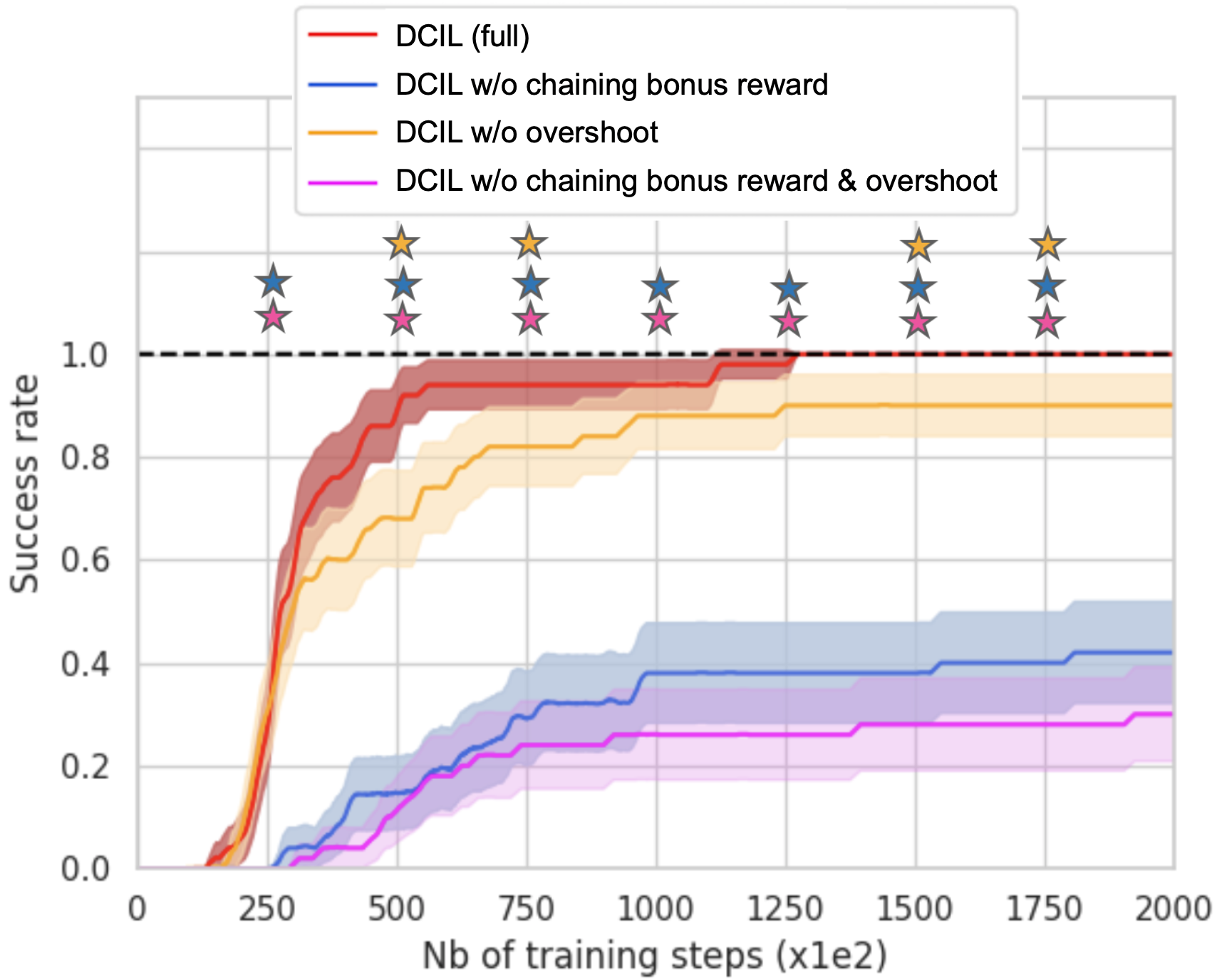}
     \hfill
     \caption{Ablation study of DCIL in the Dubins Maze environment. We evaluate the success rate of four versions of DCIL (full DCIL DCIL w/o overshoot, DCIL w/o chaining bonus reward, DCIL w/o both) throughout training. Means and standard deviations ranges over 30 total runs (3 random seeds for 10 different expert trajectories). Stars indicate significant differences over DCIL (full) as reported by Welch's t-test with $\alpha=0.05$ \cite{colas2019hitchhiker}.}
    \label{fig:sr_GGI_ablation_dubins}
\end{figure}

\begin{figure}[b!]
     \centering
     \includegraphics[width = 0.95 \hsize]{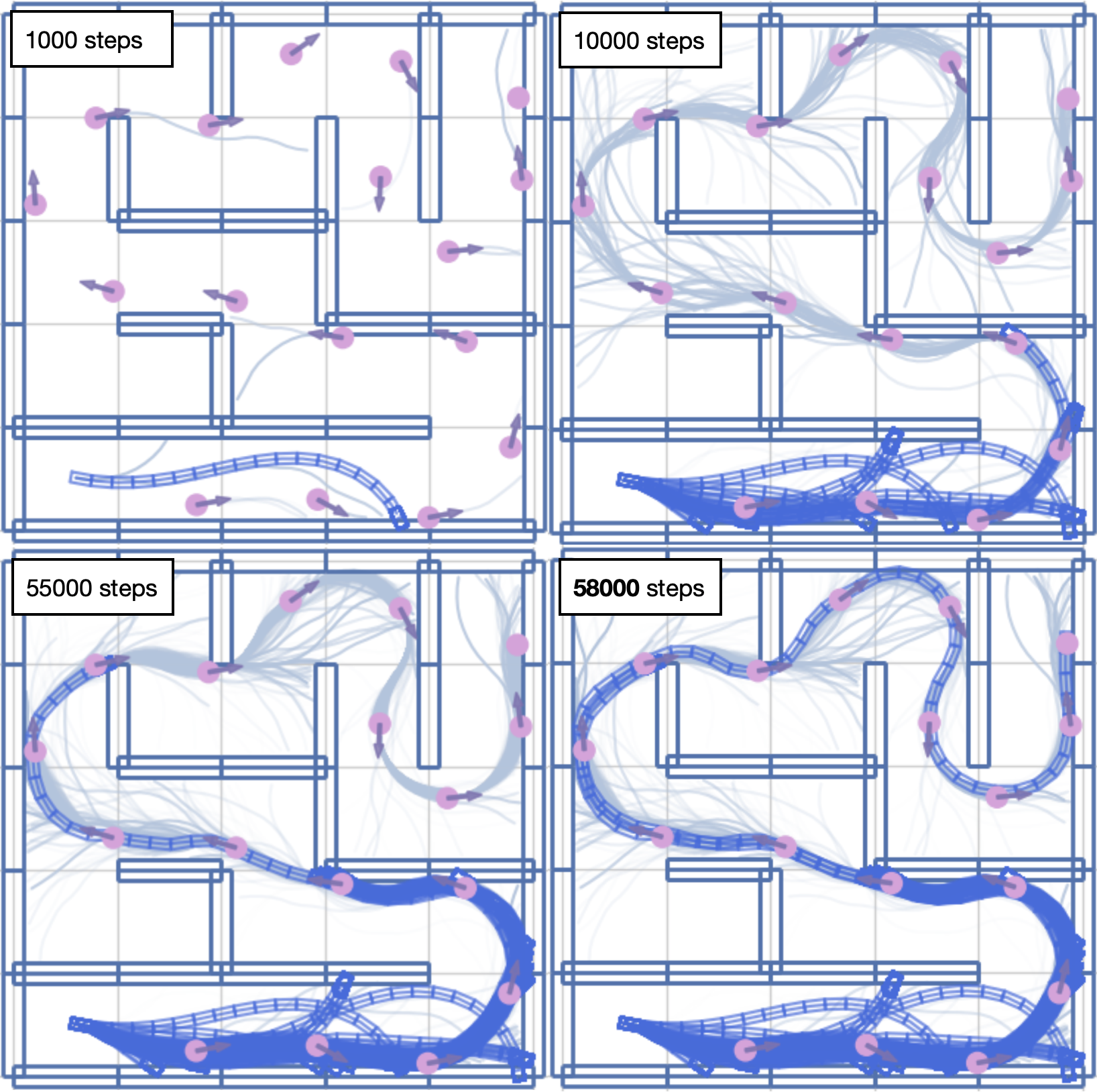}
     \hfill
     \caption{Training run of DCIL in the Dubins Maze. For each skill, the initial state is represented by a purple arrow and the success goal set by a pink disk. Grey lines correspond to skills training rollouts. Blue car trajectories correspond to the agent skill chaining every 1000 training steps.}
    \label{fig:training_GGI_dubins}
\end{figure}

\subsection{Ablation study}
\label{sec:ablation_study}

Using the Dubins Maze environment, we compare the full version of DCIL to variants without the chaining reward bonus, without the overshoot mechanism and without both.

The performance shown in \figurename~\ref{fig:sr_GGI_ablation_dubins} is evaluated using the proportion of runs that solved the maze depending on the number of training steps.
First, we can notice that the chaining reward bonus is critical to chain the skills and achieve a high success rate. Indeed, the two variants of DCIL using the chaining reward bonus (DCIL full and DCIL w/o overshoot) outperform the other two variants.  
Besides, only the full version of DCIL recovers the expert behavior in $100\%$ of trials. Finally, DCIL also benefits from the overshoot mechanism as its success rate increases faster than DCIL without overshoot during the $50.000$ first training steps.

\begin{figure}[t!]
     \centering
     \includegraphics[width = 0.9 \hsize]{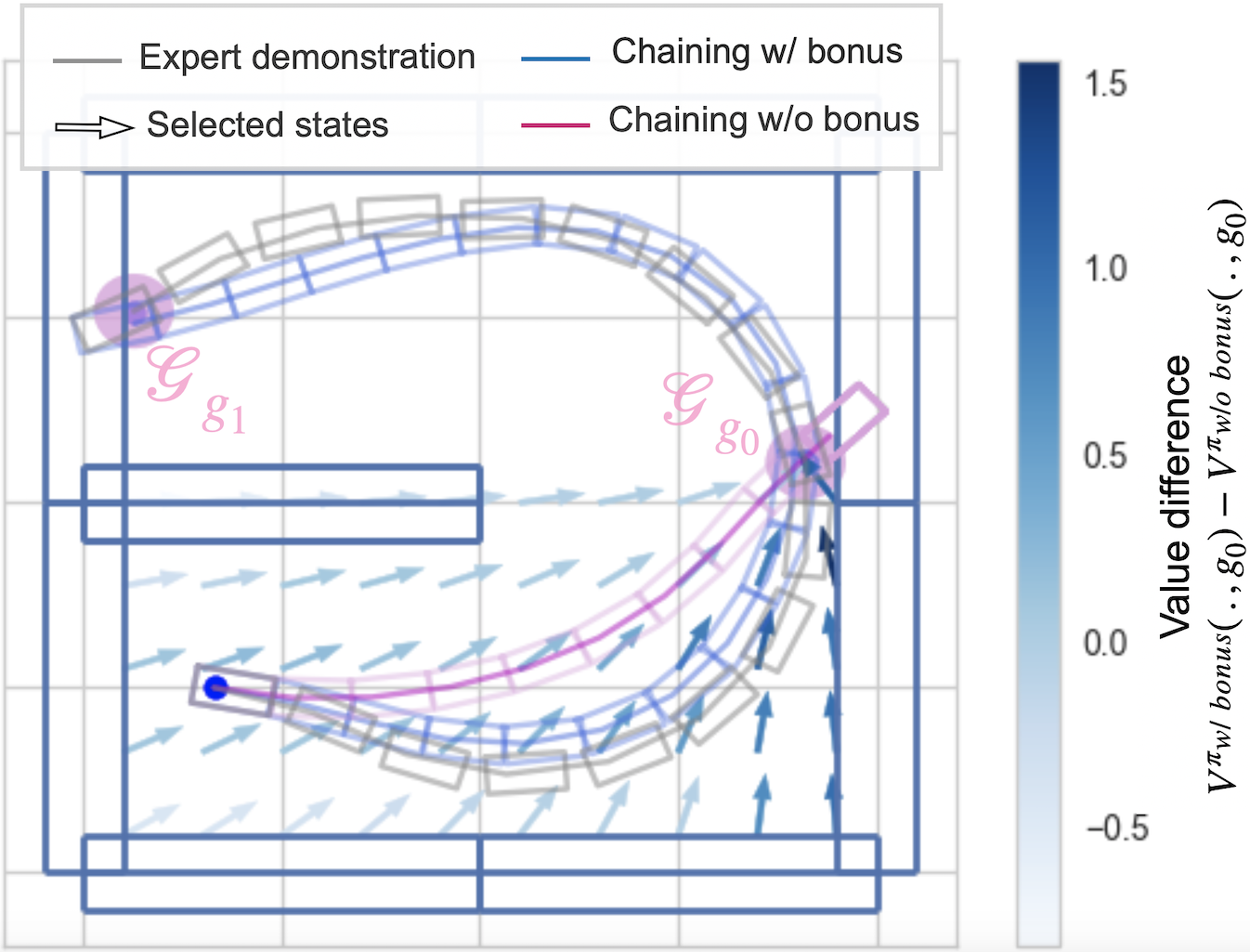}
     \hfill
     \caption{Learning the first skill (reaching $\mathcal{G}_{g_0}$) without chaining reward bonus prevents the agent from completing it by reaching an valid initial state for the second skill (reaching $\mathcal{G}_{g_1}$) as illustrated by the purple trajectory. The chaining reward bonus increases the value of the states with an orientation similar to the states of the expert demonstration (in grey) as shown by the difference between the $g_{0}$-conditioned values learned with a chaining reward bonus ($V^{\pi_{w/ bonus}}(.,g_{0})$) and without ($V^{\pi_{w/o\  bonus}}(.,g_{0})$).}
    \label{fig:toy_example_bonus_dubins}
\end{figure}

Figure~\ref{fig:training_GGI_dubins} presents a training run of the full version of DCIL. Although the GCP is training on skills separately, it is able to chain them in order to recover the expert behavior and to navigate the maze. In order to visualize how the chaining reward bonus encourages the agent to complete the skills by reaching valid initial states, we evaluated DCIL and DCIL w/o chaining bonus reward in a simplified version of the Dubins Maze where the agent only has two skills to chain. As \figurename~\ref{fig:toy_example_bonus_dubins} shows, the chaining reward bonus increases the value of the states with a similar orientation to the states in the expert demonstration and results in successful chaining of the two skills. 

\begin{figure}[b!]
     \centering
     \includegraphics[width = \hsize]{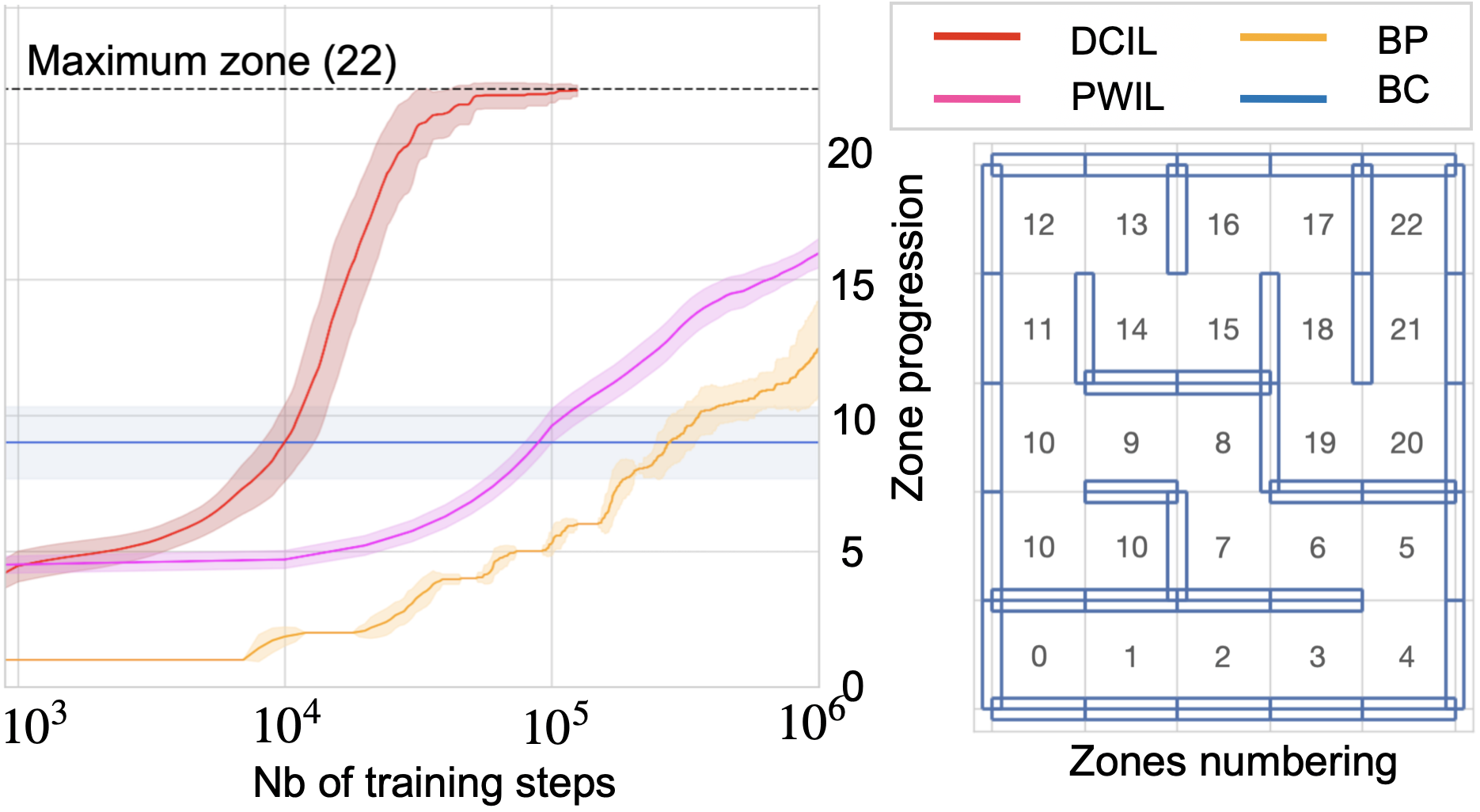}
     \hfill
    \caption{Comparison of DCIL to
    Backplay, PWIL and BC with a single demonstration. We evaluate the progression through the Dubins Maze Environment (left) using 22 zones (right). For DCIL, PWIL and BC, the progression corresponds to the maximum zone reached during evaluation. For Backplay, the progression corresponds to the difference between the highest zone number and the zone from which the agent started. Means and standard deviations ranges over 3 random seeds for 10 different expert trajectories (30 total runs for each variant) for each method.}
    \label{fig:max_zones_dubins}
\end{figure}

\subsection{Comparison to baselines in Dubins Maze}
\label{sec:baselines_dubins}

Figure~\ref{fig:max_zones_dubins} evaluates how DCIL performed when trained on $1e6$ training interactions with the Dubins Maze compared to the three selected baselines. As the three baselines do not solve the maze after $1e6$ training interactions, we use a metric based on the progression through the maze. The maze is decomposed into 23 zones. The agent starts in zone 0 and the end of the maze is zone 22. 
DCIL is the only method able to solve the maze within the allocated budget. It requires at most $10^{5}$ training interactions. This is mainly due to the fact that DCIL trains on very short rollouts compare to PWIL which trains on fixed-length episodes and Backplay which trains on episodes of increasing length.

\begin{figure}[ht]
     \centering
     \includegraphics[width = 0.8 \hsize]{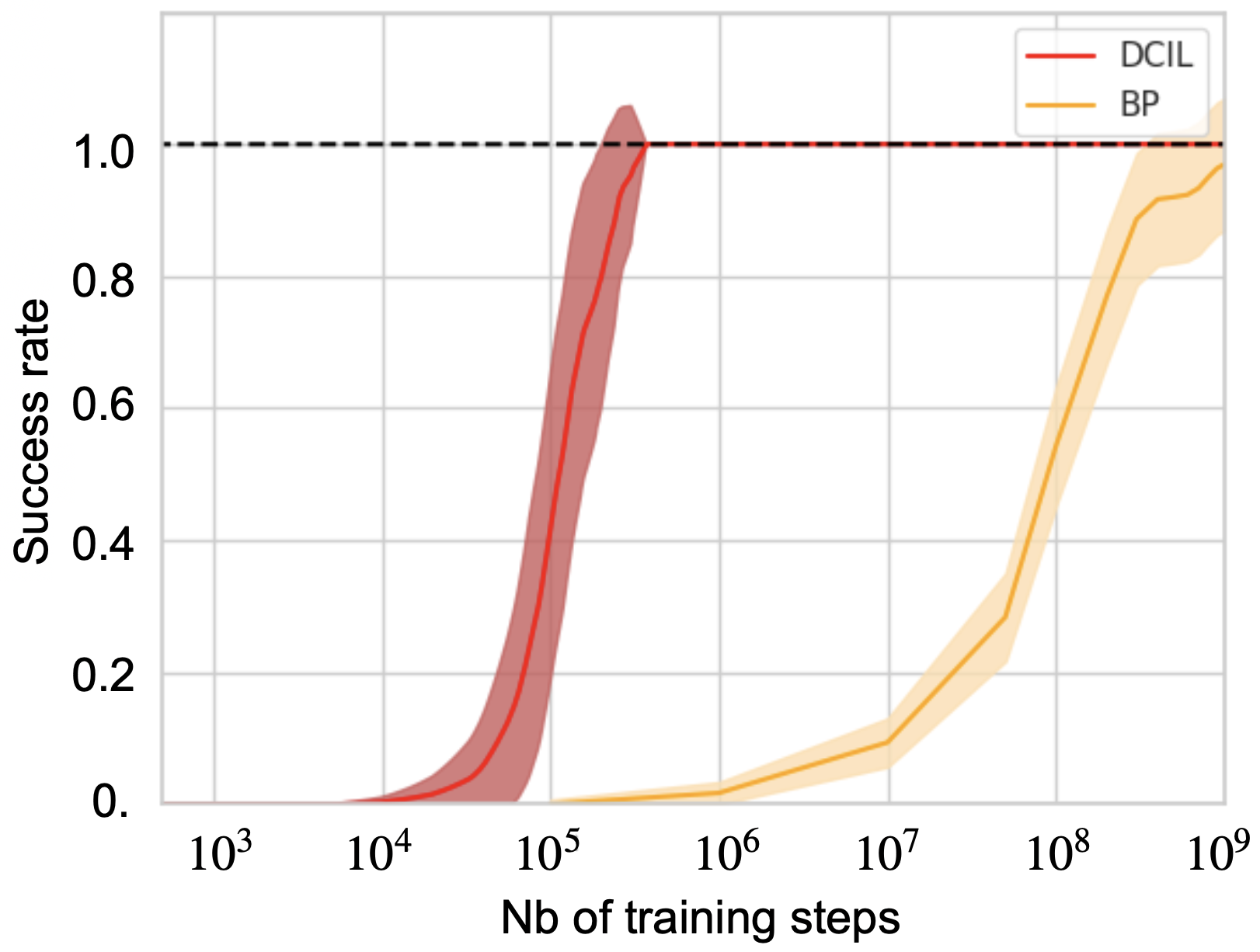}
     \hfill
    \caption{Comparison of DCIL to Backplay in the Fetch environment. The results of DCIL present the mean and standard deviation over 5 seeds for 5 expert demonstrations (25 total runs). The results of Backplay are extracted from \cite{ecoffet2021first} and could not be reproduced despite running the author's code with the same hyper-parameters. For DCIL, in $\sim10\%$ of the runs, the critic networks used in SAC diverge which results in failed runs. Only the runs that did not diverge are presented here.}
    \label{fig:sr_GGI_fetch}
\end{figure}

\subsection{Scaling to a complex object manipulation task}
\label{sec:scaling}

While the experiments in the Dubins Maze demonstrate the performance of DCIL in a low-dimensional environment, we finally test our approach in the Fetch environment where observations are 604-dimension vectors and the transition function involves a much more complicated dynamic. 

Figure~\ref{fig:sr_GGI_fetch} evaluates how DCIL performs when trained on $1e6$ training interactions. We compare our approach to the performance of Backplay presented in \cite{ecoffet2021first}. As in the Dubins Maze, DCIL solves the Fetch task three orders of magnitude faster than Backplay. DCIL is able to learn the full fetch behavior by training only on short rollouts.


\section*{Discussion \& Conclusion}
\label{sec:discussion}

In this paper, we have introduced Divide \& Conquer Imitation Learning (DCIL), an imitation learning algorithm solving long-horizon tasks using a single demonstration. DCIL relies on a sequential inductive bias and adopts a divide \& conquer strategy to learn smaller skills that, chained together, solve the long-horizon task. In order for a goal-conditioned policy to learn each skill individually and to apply skill-chaining to recover the expert behavior, we introduced an \textit{overshoot mechanism} and a \textit{chaining reward bonus} that indirectly make skills aware of the next ones, and significantly improves the chainability of the skills. We highlighted the key contribution of both mechanisms in the performance of DCIL by conducting an ablation study in a maze environment with a Dubins car. Moreover, we showed the efficiency of DCIL by comparing it to three IL baselines and by successfully applying it to a complex manipulation task. 

Compared to the baselines, we obtain an improvement of sample efficiency of several orders of magnitudes, which, in future work, will be critical when applying the method to physical robots. Yet, the application of DCIL to physical robots would require at least one modification of the algorithm which concerns the reset assumption. The usual option would be to replace resets to any previously encountered state by a unique fixed reset. This could be done by learning a way to "return to interesting states", an approach that has been studied in one of the variants of the Go-Explore algorithm \cite{ecoffet2021first}. However, even this assumption of a single reset can be troublesome with physical robots, especially when object manipulation is involved. For this reason, our main research direction will be to consider the possibility of learning small robotic skills that are not only chainable but \emph{reversible}, which in particular robotic contexts could lead to a sample-efficient divide \& conquer approach for imitation learning that would not require any kind of reset at all.


\section*{Acknowledgements}

This work was partially supported by the French National Research Agency (ANR), Project ANR-18-CE33-0005 HUSKI and was performed using HPC resources from GENCI-IDRIS (Grant 2022-A0111013011) and the MeSU platform at Sorbonne Université. We want to thank Ahmed Akakzia and Elias Hanna for the fruitful discussions.

\bibliographystyle{IEEEtran}

\newpage

\end{document}